# Pragmatic inference of scalar implicature by LLMs

**Ye-eun Cho** and **Seong mook Kim**
Department of English language and literature
Sungkyunkwan University, South Korea
`{joyenn, ismkim99}@skku.edu`

## Abstract

This study investigates how Large Language Models (LLMs), particularly BERT (Devlin et al., 2019) and GPT-2 (Radford et al., 2019), engage in pragmatic inference of scalar implicature, such as *some*. Two sets of experiments were conducted using cosine similarity and next sentence/token prediction as experimental methods. The results in experiment 1 showed that, both models interpret *some* as pragmatic implicature *not all* in the absence of context, aligning with human language processing. In experiment 2, in which Question Under Discussion (QUD) was presented as a contextual cue, BERT showed consistent performance regardless of types of QUDs, while GPT-2 encountered processing difficulties since a certain type of QUD required pragmatic inference for implicature. The findings revealed that, in terms of theoretical approaches, BERT inherently incorporates pragmatic implicature *not all* within the term *some*, adhering to Default model (Levinson, 2000). In contrast, GPT-2 seems to encounter processing difficulties in inferring pragmatic implicature within context, consistent with Context-driven model (Sperber and Wilson, 2002).

## 1 Introduction

In recent years, there has been remarkable progress in Natural Language Processing (NLP) thanks to the advent of Transformers (Vaswani et al., 2017), from which numerous Large Language Models (LLMs) have been developed. The effectiveness of these models relies on their ability to comprehend user input, which demands a focus on both semantics and pragmatics. Semantics involves the literal meanings of words or sentences, while pragmatics focuses on context-dependent intended meanings. Although advances in language modeling, particularly in neural vector representations like Word2Vec (Mikolov et al., 2013) and GloVe (Pennington et al., 2014), have shown significant progress in semantics, pragmatic inference has not received as much attention in NLP research, despite its importance for achieving increasingly natural conversations with users.

Pragmatic inference refers to the process of making inference by considering the contexts, intentions, and situations of language use. As a type of pragmatic inference, **implicature** is regarded as a linguistic phenomenon where the speaker conveys additional meaning or information that is not explicitly stated. One of the most commonly studied implicatures is **scalar implicature**, which indicates the quantity or range of a particular attribute, such as *some*. Logically and semantically, the term *some* means *at least one and possibly all*. But, in actual language use, *some* is not always interpreted in this manner. Pragmatically, *some* would lead the hearer to infer the meaning *not all*.

(1)  Some students passed the exam.

For example, the sentence in (1) might be recognized as not all students passed the exam rather than at least one (or two in this case) and possibly all of them did.

However, Roberts (2012) suggested that, in pragmatic discourse, whether *some* is interpreted semantically or pragmatically depends on the surrounding context, such as **Question Under Discussion (QUD)**. QUD refers to topics in a conversation that should be addressed with relevant responses at a later stage in communicative interaction (Roberts, 2004; 2012; Beaver and Clark, 2008).

(2)  A: Did all students pass the exam?
     B: Some students passed the exam.

Considering a conversational exchange in the form of QUD as in (2), *some* is more clearly interpreted as *not all* due to A's question. This illustrates that *some* and *all* are positioned together or mutually related on an informational scale as <*some, all*>, on which the less informative or weaker term *some* implies the negation of the more informative or stronger term *all* (Horn, 1972).

Several studies have attempted to explore whether LLMs can learn scalar implicature through Natural Language Inference (NLI) tasks (Jeretic et al., 2020; Schuster et al., 2020; Li et al., 2021). However, to our knowledge, the effects of manipulating context on scalar implicature have not been explored. Therefore, this study aims to investigate whether LLMs lean towards a semantic or pragmatic interpretation of scalar implicature and whether the interpretation can be influenced by context, drawing insights from experiments conducted in human language processing.

## 2 Background

### 2.1 Interpretations of scalar implicature

The study of deriving scalar implicature for the quantifier *some* has been widely conducted to investigate how pragmatically enriched meanings are computed. For example, the utterance in (3) semantically entails that *at least two and possibly all students passed the exam*, while pragmatically it is interpreted as *not all students passed the exam*, in which the meaning is enriched by the implicature (Geurts and Nouwen, 2007; Cummins and Katsos, 2010; Geurts et al., 2010).

(3) a. *Utterance*:
    Some students passed the exam.
  b. *Semantic entailment*:
    At least two and possibly all students passed the exam.
  c. *Pragmatic implicature*:
    Not all students passed the exam.

These two interpretations differ in whether *all* is negated or not, allowing for the possibility that *all* may still be valid in semantic interpretation. Furthermore, as shown in (4), the semantic entailment *at least one and possibly all* is not cancellable, while the interpretation of pragmatic implicature *not all* is cancellable (Grice, 1989; Geurts, 2010).

(4) a. *Non-cancellable semantic entailment*:
    Some students passed the exam. #In fact, none of them did.
  b. *Cancellable pragmatic implicature*:
    Some students passed the exam. In fact, all of them did.

This leads to the argument that *some* is positioned on a quantifier scale with varying levels of informativeness, ranging from the least to the most informative, representing the continuum <*some, all*> (Horn, 1972). The informativeness on the quantifier scale corresponds to the scale strength, where the less informative items are relatively weaker while the more informative ones are relatively stronger on the scale.

It is also argued that the hearer generally infers the speaker's intention not to use the strong item (i.e., *all*) when trying to convey the meaning of the weak item (i.e., *some*). This is because interlocutors in conversation often expect that the speaker's utterance would be optimally informative, as generalized by Gricean maxims (Grice, 1975). Therefore, scalar implicature leads to the general perception that the weak term implies the negation of the strong term on the scale.

However, *some* is not always interpreted with *not all* implicature. Roberts (2004) and Chierchia et al. (2012) have shown that the interpretation of *some* is heavily dependent on the broader context. Specifically, Roberts (2012) argued that whether *some* is interpreted with the pragmatic implicature is determined by the QUD, which refers to the topics in conversation that are expected to be addressed by appropriate answers (Roberts, 2004; 2012; Beaver and Clark, 2008). Examples can be found in (5) and (6), where the utterances containing *some* occur in response to different questions. The QUD that contains the term *all* is regarded as upper-bound as in (5), while the QUD that contain *any* is regarded as lower-bound as in (6).

(5) *Upper-bound QUD*:
    A: Did all students pass the exam?
    B: Some students passed the exam.

(6) *Lower-bound QUD*:
    A: Did any students pass the exam?
    B: Some students passed the exam.

In the upper-bound QUD, the utterance of B is clearly interpreted as *not all students passed the exam*, suggesting *not all* implicature. On the other hand, the utterance of B in the lower-bound QUD can be felicitously interpreted, without *not all* implicature, as *at least two and possibly all students passed the exam*. The distinct interpretations of the same utterance in (5) and (6) arise due to the different questions asked by the speaker A. This illustrates that the utterance containing *some* may be ambiguous without any context, whereas a contextual cue, such as the QUD, can disambiguate the optimal interpretation of *some* in the discourse.

## 2.2 The processing of scalar implicature

Many studies have experimentally investigated whether scalar implicature is interpreted in semantic or pragmatic manner. For example, Bott and Noveck (2004) asked participants to judge the sentence in (7) is true or false.

(7) Some elephants are mammals.

Based on world knowledge, if *some* was interpreted semantically as *at least one and possibly all*, this sentence would be true; however, if *some* was interpreted pragmatically as *not all*, this sentence would be false. As a result, more participants judged these kinds of sentences as false, indicating a preference for pragmatic interpretation rather than semantic interpretation when scalar implicature was presented without context. These results have consistently appeared in other studies (Noveck and Posada, 2003; De Neys and Schaeken, 2007; Huang and Snedeker, 2009; Hunt et al., 2013; Tomlinson et al., 2013).

There have been two approaches to explain the processing of scalar implicature: Default model (Levinson, 2000) and Context-driven model (Wilson and Sperber, 1995; Sperber and Wilson, 2002). Levinson (2000) suggested, from the perspective of the Default model, that the hearer generally has an expectation of how language is typically used. This leads to *not all* implicature by default when encountering the term *some*. That is, implicature is generated as a default and can be negated or canceled when it becomes irrelevant in the given context. In contrast, Sperber and Wilson (2002) argued that scalar implicature is processed based on Relevance Theory. According to Relevance Theory, human cognition is generally inclined to maximize relevance (Wilson and Sperber, 1995). This inclination allows a given utterance to be integrated with context, resulting in more positive cognitive effects for a more relevant utterance, while requiring greater processing effort for a less relevant utterance. In this view, the context plays a crucial role in determining whether the implicature is generated in the first place.

To examine the impact of context in the processing of scalar implicature, several studies have incorporated QUD in their experiments (Breheny et al., 2006; Zondervan et al., 2008; Politzer-Ahles and Fiorentino, 2013; Degen and Goodman, 2014; Dupuy et al., 2016; Politzer-Ahles and Husband, 2018; Yang et al., 2018; Ronai and Xiang, 2020).

(8) A: Did you fold all/any sweaters?
B: I folded some sweaters.

For example, Yang et al. (2018) presented participants with a situation where sentences containing *some* were followed by questions, as in (8). The QUD including *all* in the question is relevant to pragmatic implicature (i.e., *not all*), whereas the QUD including *any* in the question does not require implicature to interpret the conversation. In terms of pragmatic implicature, weak item *some* carries the meaning of negating the strong item *all*. Thus, if pragmatic implicature is appropriately established, the ratings for the sentences containing *some* should be lower, and the cognitive efforts required to infer the implicature should be greater in the *all*-condition than those in the *any*-condition. The experimental results exhibited that *all*-condition was rated lower than *any*-condition, suggesting that the interpretations of scalar implicature are sensitive to the given context. In addition, cognitive efforts measured in this study were greater when interpreting *some* in the upper-bound QUD (i.e., *all*-condition). This finding supports Context-driven model (Wilson and Sperber, 1995; Sperber and Wilson, 2002), indicating that more cognitive effort is required to derive scalar implicature.

Drawing from studies of human language processing related to scalar implicature, the current study poses the following questions regarding the language processing abilities of LLMs:

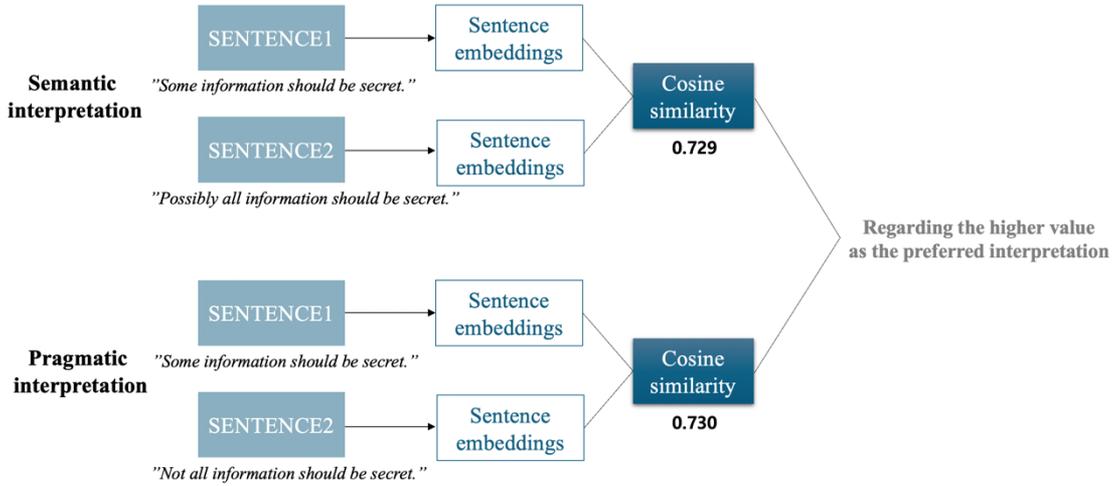

Figure 1. Overview of embedding *some*-sentences and its semantic and pragmatic counterparts, measuring cosine similarities between SENTENCE1 and SENTENCE2, and selecting the models' preferred interpretation between semantic and pragmatic interpretations in experiment 1

1. Do LLMs perform pragmatic interpretation rather than semantic interpretation for scalar implicature without context?
2. Do LLMs exhibit sensitivity to a contextual cue, such as QUD, in discourse during the processing of scalar implicature?

To address these questions, we will conduct two experiments in the following sections.

## 3 Data collection

To investigate the processing of scalar implicature by LLMs, we extracted sentences with '*some* + NP' structures from British National Corpus (BNC) using NLTK (Bird, 2006). Among those, we collected sentences where '*some* + NP' was positioned as the subject due to the fact that the implicature generation is stronger when '*some* + NP' is positioned at the sentence-initial position compared to the sentence-final position (Breheny 2006). In addition, we excluded sentences with multiple clauses to avoid the possibility of cancellation. Finally, a total of 198 sentences were extracted and one example of the final data is presented as in (9).

(9) Some information should be secret.
(BNC W:newsp:other:social, K5C-156)

| SENTENCE1 | SENTENCE2 | Interpretation |
|---|---|---|
| **Some** information should be secret. | **Possibly all** information should be secret. | Semantic |
| **Some** information should be secret. | **Not all** information should be secret. | Pragmatic |

Table 1. Materials for experiment 1

We refer to the sentences extracted through this process as *some*-sentences. Both data and results of the experiments are publicly available.[1]

## 4 Experiment 1

Previous experiments on human language processing have successfully captured pragmatic inference of scalar implicature even without context (Noveck and Posada, 2003; De Neys and Schaeken, 2007; Huang and Snedeker, 2009; Hunt et al., 2013; Tomlinson et al., 2013). Likewise, experiment 1 aimed to investigate how LLMs interpret *some*-sentences without context, distinguishing between semantic entailment or pragmatic implicature.

### 4.1 Method

The experimental materials consisted of *some*-sentences and sentences with its semantic and pragmatic interpretations as shown in Table 1.
SENTENCE1 was composed of the *some*-sentences, while SENTENCE2 included sentences with either semantic or pragmatic interpretations. To ensure uniform token count between the two sentences in SENTENCE2, the sentence with the semantic interpretation used only *possibly all* instead of *at least one and possibly all*. Each pair of SENTENCE1 and SENTENCE2 was labeled as either 'Semantic' or 'Pragmatic' depending on its interpretation.

---

[1] https://github.com/joyennn/scalar-implicature

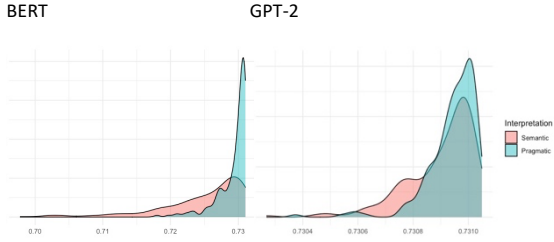

Figure 2. Density of cosine similarities between *some*-sentences and its semantic or pragmatic interpretations

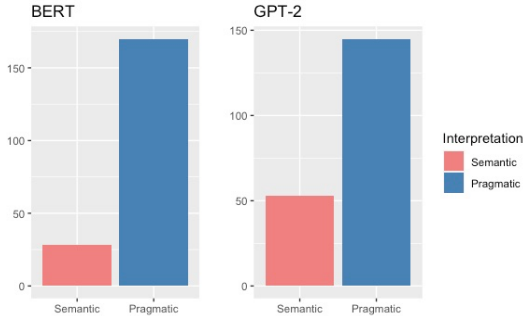

Figure 3. Instances with higher similarities for the same *some*-sentences across interpretations

LLMs used for the experiment were BERT (Devlin et al., 2019) and GPT-2 (Radford et al., 2019), both of which were transformers-based pre-trained language models. BERT (`BERT-base-uncased`) comprises 12 Transformer encoder layers, each of which is designed to capture bidirectional context from the input text. Unlike BERT, which uses only encoder layers, GPT-2 small (`gpt2`) utilizes 12 decoder layers to generate text in an autoregressive manner, predicting the next word in a sequence based on the previously generated words. Despite these differences, both models have a hidden size of 768 and 12 self-attention heads. In addition, the total number of parameters are similar in BERT and GPT-2 which have approximately 110 million and 117 million parameters, respectively. Although newer and more advanced models have proved higher performance, BERT and GPT-2, as foundational transformer models, are well-known in terms of their processing architectures, which allows us to better understand how these models process language.

Specifically, input sentences were tokenized using each model's tokenizer. For BERT, we obtained sentence embeddings by using the [CLS] token embeddings from the final layer. On the other hand, for GPT-2, sentence embeddings were derived by averaging the token embeddings from the final layer. We then computed the cosine similarity between pairs of corresponding sentence embeddings for SENTENCE1 and SENTENCE2. Cosine similarity is a method that measures how similar two sentences are by evaluating the angle between two sentence vectors. Although it may underestimate the similarity of words or sentences (Zhou et al., 2022), it is not just suitable for measuring the similarity of sentences but also computationally efficient and widely used in many studies. These cosine similarity scores were averaged to obtain a single similarity measure for the sentence pairs.

Since the value of cosine similarity ranges from [-1, 1], it was linearly transformed to a [0, 1] range for ease of interpretation. Then, the sigmoid function was applied to ensure to avoid values that are extremely close to 0 or 1. In this classification, a value close to 1 indicates high similarity between two sentences, while a value close to 0 indicates lower similarity. Through this metric, we measured whether the *some*-sentences were interpreted in a semantic or pragmatic manner. The overview of the experiment 1 is presented in Figure 1.

To verify statistical significance, a linear mixed-effects regression model from the lme4 package in the R statistical software was employed (Bates et al. 2014). The summaries of linear mixed-effects models are provided in the Appendix section.

### 4.2 Result

Figure 2 showed the density of the similarities between *some*-sentences and its semantically or pragmatically interpreted counterparts. While both interpretations exhibited similarities between 0.5 and 1, indicating high degree of sentence similarities, the pragmatic interpretations appeared relatively more prominent.

Figure 3 illustrated which interpretations, semantic or pragmatic, exhibited higher similarities for the same *some*-sentences. In BERT, 28 instances showed higher similarities to the semantic interpretations while 170 instances showed higher similarities to the pragmatic

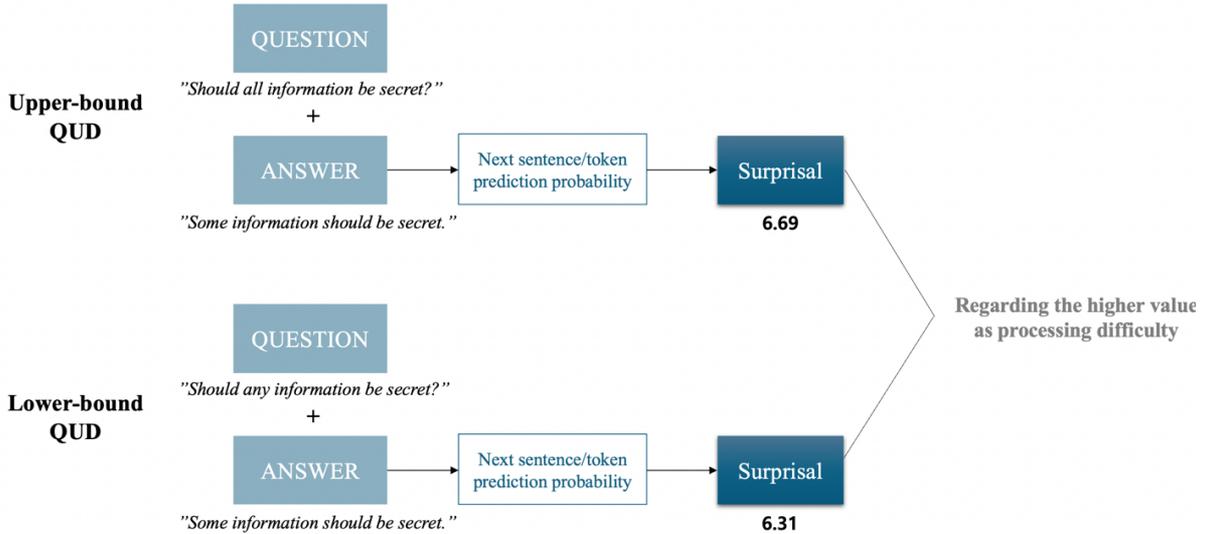

Figure 4. Overview of embedding question and answer sentences, calculating next sentence/token prediction probabilities for the answer sentences, and transforming the probabilities into surprisals in experiment 2

interpretations. In GPT-2, 53 instances exhibited higher similarities to the semantic interpretations while 145 instances exhibited higher similarities to the pragmatic interpretations. The statistical analysis revealed significant effects in the interpretations for both models ($p < 0.001$).

In summary, the interpretations of the scalar implicature *some* without context tended to be predominantly pragmatic, reflecting a consistency with human language processing.

## 5 Experiment 2

Building on the findings from Experiment 1, which showed that both BERT and GPT-2 models prefer pragmatic interpretations to semantic interpretations in scalar implicature without context, experiment 2 aimed to explore whether the LLMs have more processing difficulties when implicature is required (i.e., upper-bound QUD), compared to when implicature is not required (i.e., lower-bound QUD). For this comparison, the context was manipulated using QUD as a contextual cue.

### 5.1 Method

In the experimental materials, two types of the question sentences were generated for the *some*-sentences according to the types of QUDs, such as upper- and lower-bound QUDs. Following Yang et al. (2018), questions for the upper-bound included *all*, while those for the lower-bound included *any*.

| QUESTION | ANSWER | QUD |
|---|---|---|
| Should **all** information be secret? | **Some** information should be secret. | Upper |
| Should **any** information be secret? | **Some** information should be secret. | Lower |

Table 2. Materials for experiment 2

As presented in Table 2, QUESTION comprised questions with either *all* or *any*, while ANSWER consisted of the *some*-sentences. Each pair of QUESTION and ANSWER was labeled as either 'Upper' or 'Lower' depending on its QUD.

In experiment 2, we also employed BERT-base and GPT-2 small models. BERT is pre-trained using Next Sentence Prediction (NSP), which involves predicting whether the second sentence immediately follows the first sentence in the given pair of sentences. This is achieved by concatenating the two sentences with [CLS] (classification start) and [SEP] (sentence separator) tokens to form the input for the BERT model. The [CLS] token embeddings from the final layer are used to compute the NSP probability, thereby quantifying the probability of ANSWER following QUESTION.

On the other hand, GPT-2 does not utilize methods like BERT's NSP as its training data lacks explicit signals indicating relationships between sentences. Instead, GPT-2 predicts the next word based on the preceding context. To assess the relationship between two sentences in GPT-2, we combined QUESTION and ANSWER into a single text sequence. After providing this combined sequence as input to GPT-2, we analyzed the probability of the next generated token. This probability was used to estimate the likelihood of ANSWER appearing after QUESTION.

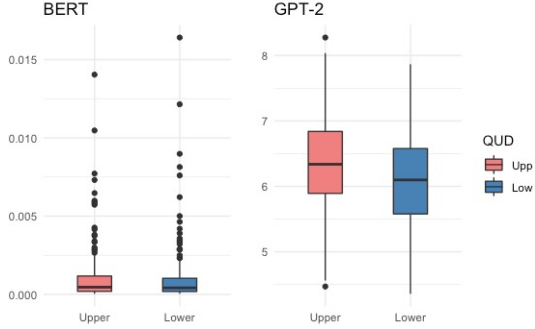

Figure 5. Distribution of surprisal scores for processing *some*-sentences across QUDs

The output probabilities ($P$) from both models were transformed into **Surprisal** (Hale, 2001; Levy, 2008). Surprisal plays an effective role in measuring cognitive effort in human language processing. In this case, surprisal was used to measure models' processing difficulties. As shown in (10), this value is inversely correlated with how acceptable the next sentence ($S$) is in the given context (*Context*).

(10) $Surprisal = -log_2 P\ (S|Context)$

With surprisal scores, we could compare the processing difficulties of the models in the upper- and lower-bound QUDs. The overview of the experiment 1 is shown in Figure 4.

### 5.2 Result

Figure 5 depicted the distribution of surprisal scores for each model across QUDs. BERT showed little difference in surprisals based on QUDs (median of Upper = 0.00045, median of Lower = 0.00041), and statistically, no main effects were observed ($p = 0.48$). This suggested that BERT was unaffected by context in the interpretation of scalar implicature. Conversely, GPT-2 exhibited higher surprisal scores for the upper-bound QUD (median = 6.33) compared to the lower-bound QUD (median = 6.09), and this result was statistically significant ($p < 0.01$). This processing pattern of GPT-2 was consistent with human language processing, which suggested that GPT-2 showed processing difficulties, similar to the greater cognitive effort that humans expend in inferring scalar implicature in the context of QUD.

In summary, while exploring the interpretation of scalar implicature across QUDs, BERT exhibited no sensitivity to context, whereas GPT-2 clearly manifested the effects of context.

## 6 Discussion

Through two sets of experiments, this study investigated how LLMs interpret scalar implicature, between semantic entailment and pragmatic implicature, in the absence of context and how QUD, as a contextual cue, affects LLMs' processing of scalar implicature. Experiment 1 investigated whether *some*-sentences in BERT and GPT-2 exhibit greater similarity to semantic or pragmatic interpretations. The results showed that both models preferred the interpretation of pragmatic implicature over semantic entailment for *some*-sentences. Experiment 2 aimed to investigate whether providing QUD as a contextual cue would impact processing difficulties for BERT and GPT-2, comparing between the upper-bound QUD, where pragmatic implicature is required, and the lower-bound QUD, where implicature is not required. As a result, BERT showed no significant difference in processing difficulties based on QUDs, whereas GPT-2 showed more processing difficulties in the upper-bound QUD. In conclusion, this study found that, only in a certain language model, GPT-2, greater processing difficulties were captured during pragmatic inference of scalar implicature, aligning with human language processing.

BERT and GPT-2, despite both being built on the transformer architecture, exhibited markedly different patterns regarding their theoretical approaches to the processing of scalar implicature. Although both models shared the patterns of interpreting the term *some* as a pragmatic *not all* implicature rather than a semantic *at least one and possibly all* without context, BERT exhibited no discernible difference in processing based on QUDs. This can be explained by Default model where the meaning of *some* inherently defaults to *not all* (Levinson, 2000). On the other hand, GPT-2 represented a clear difference in processing difficulties when manipulating context through the setting of QUD, revealing that greater processing difficulties were captured in the processing of scalar implicature. This finding follows Context-driven model, consistent with the argument that *not* *all* implicature is not inherently embedded to the term *some* but rather inferred through a broader context (Wilson and Sperber, 1995; Sperber and Wilson, 2002).

Among the earlier NLI studies regarding scalar implicature, Jeretic et al. (2020) found that BERT learned scalar implicature. They claimed that positive results on scalar implicature inference, triggered by specific lexical items like *some* and *all*, probably exploits prior knowledge during the pre-training stage. The natural language data employed in the pre-training inherently include pragmatic information, which raises the possibility that such pre-training induces patterns of pragmatic inference in the data. Therefore, the results of the experiment 1 in this study, where the interpretation of pragmatic implicature occurred even in the absence of context, can be explained as leveraging inherent pragmatic information in the pre-training data of LLMs.

In the study of Schuster et al. (2020), which investigated the effects of linguistic features on scalar implicature, they found that their model could make accurate predictions without considering the preceding context, while incorporating the preceding conversational context did not enhance and even diminished prediction accuracy. This led to the assumptions that only a context-independent utterance is sufficient and contextual cues may not be necessary for pragmatic inference, or that the model has not appropriately used contextual information. Finding that context is unnecessary in scalar implicature may provide an explanation for our observation that BERT in the experiment 2 showed no difference in processing efforts across QUDs. However, this explanation may not generalize to effectively capture the processing of scalar implicature in all LLMs, especially when taking into account the effects of QUD on the processing of scalar implicature in GPT-2.

Liu et al. (2019) reported that features generated by pre-trained contextualizers were sufficient for achieving high performance across a broad range of tasks which explored the linguistic knowledge and transferability of contextualized word representations. However, they proposed that, for tasks requiring specific information not captured by contextual word representations, learning task-specific contextual features plays a crucial role in encoding the requisite knowledge. Within this framework, pragmatic implicature may either be pre-trained or require additional learning processes, depending on LLMs. Therefore, it is crucial to recognize that different language models may incorporate diverse linguistic information and exhibit distinct processing patterns for the same linguistic phenomenon.

Furthermore, based on the argument of Degen and Tanenhaus (2015, 2016) in which humans are influenced by context-driven expectations about unspoken alternatives, Hu et al. (2023) examined the BERT model's variation in scalar implicature rate not just within a single scale like <*some, all*> but also across scales with diverse lexical items as unspoken alternatives of *some*. This study revealed that the model's ability to make pragmatic inferences becomes stronger as more alternatives become available, which is depending on contextual predictability. This result leads us to expect that BERT will show contrasting result if more alternatives are presented and the context becomes more predictable, despite the failure to make pragmatic inference within the provided context in the present study.

In conclusion, the findings of this study suggested that LLMs are capable of pragmatic inference for scalar implicature without context. However, it is essential to understand the degree of contextual information utilization in each model and ensure appropriate learning for specific tasks.

# 7 Limitations

While this study has advanced our comprehension of pragmatic inference in LLMs regarding scalar implicature, it faces limitations in three aspects.

The first limitation is the absence of diverse constructions in which the scalar quantifier *some* appears. The exclusive use of experimental sentences featuring '*some* + NP' in the subject position within a single clause may not fully capture the broad spectrum of pragmatic interpretations that arise in various linguistic constructions and meanings in the real world. Additionally, the number of data used in the experiments might be not large enough to generalize the findings.

Secondly, the study relies on only two of early transformer-based models, which may not reflect the performance of more advanced models that have emerged recently. Since newer and more advanced models have demonstrated significantly higher performance across a wide range of tasks, using different models could yield varying results.

Lastly, in order to draw comparisons with human language processing, the experimental designs in this study deviate from conventional Natural Language Inference (NLI) tasks. Moreover, the metrics used in this study (i.e., cosine similarity or next sentence prediction) may yield different results when other metrics are applied. The diversity in experimental methodologies can lead to variations in results, emphasizing the necessity for future research to take into account such differences.

## 8 Conclusion

In this study, we discovered that LLMs interpret scalar implicature through pragmatic rather than semantic interpretation. Additionally, the study identified the model that engage in pragmatic inference through the processing of scalar implicature using a contextual cue, such as QUD, in contrast to the model that do not employ pragmatic inference. This study not only contributes to our comprehension of how LLMs process complex linguistic phenomena but also underscores the importance of considering pragmatics in NLP. By shedding light on the interplay between context and pragmatic inference, this study advances our understanding of LLMs and provides valuable insights for refining language models and applications in NLP.

## Acknowledgement

We would like to thank Professor Hanjung Lee for her valuable advice and guidance on the development of the theoretical aspects of this study. We also extend our sincere thanks to the anonymous reviewers for their insightful comments and suggestions.

# A  Appendix

|  | Estimate | Std | t | *p*-value |
|---|---|---|---|---|
| (Intercept) | 7.29E-01 | 3.44E-04 | 2121.2 | **<0.001** |
| Interpretation | -4.39E-03 | 4.02E-04 | -10.92 | **<0.001** |

Table 3. Summary of fixed effects from linear mixed-effects models by BERT in experiment 1

|  | Estimate | Std | t | *p*-value |
|---|---|---|---|---|
| (Intercept) | 7.309e-01 | 7.837e-06 | 93263.7 | **<0.001** |
| Interpretation | 4.267e-05 | 7.731e-06 | 5.519 | **<0.001** |

Table 4. Summary of fixed effects from linear mixed-effects models by GPT-2 in experiment 1

|  | Estimate | Std | t | *p*-value |
|---|---|---|---|---|
| (Intercept) | 1.077e-03 | 1.311e-04 | 8.21 | **<0.01** |
| QUD | -3.487e-05 | 4.949e-05 | -0.70 | 0.48 |

Table 5. Summary of fixed effects from linear mixed-effects models by BERT in experiment 2

|  | Estimate | Std | t | *p*-value |
|---|---|---|---|---|
| (Intercept) | 6.35 | 0.05 | 123.85 | **<0.01** |
| QUD | -0.25 | 0.01 | -17.68 | **<0.01** |

Table 6. Summary of fixed effects from linear mixed-effects models by GPT-2 in experiment 2